\pdfoutput=1

\documentclass[11pt]{article}
\usepackage[dvipsnames]{xcolor}
\usepackage{emnlp2021}

\usepackage{times}
\usepackage{latexsym}
\usepackage{graphicx}
\usepackage[T1]{fontenc}

\usepackage[utf8]{inputenc}

\usepackage{microtype}

\title{Learned Construction Grammars Converge Across Registers \\ Given Increased Exposure}

\author{Jonathan Dunn \\
  University of Canterbury \\
  Christchurch, NZ \\
\texttt{first.last@canterbury.ac.nz} \\
   \And
  Harish Tayyar Madabushi \\
  University of Sheffield \\
  Sheffield, UK \\
  \texttt{f.last@sheffield.ac.uk} \\
  }

\begin{document}
\maketitle
\begin{abstract}
This paper measures the impact of increased exposure on whether learned construction grammars converge onto shared representations when trained on data from different registers. Register influences the frequency of constructions, with some structures common in formal but not informal usage. We expect that a grammar induction algorithm exposed to different registers will acquire different constructions. To what degree does increased exposure lead to the convergence of register-specific grammars? The experiments in this paper simulate language learning in 12 languages (half Germanic and half Romance) with corpora representing three registers (Twitter, Wikipedia, Web). These simulations are repeated with increasing amounts of exposure, from 100k to 2 million words, to measure the impact of exposure on the convergence of grammars. The results show that increased exposure does lead to converging grammars across all languages. In addition, a shared core of register-universal constructions remains constant across increasing amounts of exposure.
\end{abstract}

\section{Exposure and Convergence}

The central question that this work aims to answer is whether register-specific grammars  \textit{converge} onto shared representations when exposed to more training data. Variation in the context of production, called register variation, has a significant impact on the frequency of constructions. For example, imperative and \textsc{wh}-question constructions are much more frequent in informal or conversational speech, while declarative constructions are much more frequent in formal written usage~\cite{fodor2002understanding,Sampson2002}. 

At the same time, \textit{usage-based grammar} views language as a complex adaptive system that emerges given exposure to usage \cite{Bybee2006,Beckner2009}. Thus, a language learner is expected to be strongly influenced by the observed frequency of constructions. Given this wide variance in the frequency of constructions across registers, it is conceivable that learners exposed to different registers learn different constructions.

This paper simulates the language acquisition process from a usage-based perspective by learning Construction Grammars, called CxGs~\cite{goldberg1995constructions,g06a,l08}. A constructional approach to language focuses on symbolic form-meaning mappings that are potentially idiomatic. Previous work on computational CxG has explored how to discover potential constructions \cite{Wible2010,Forsberg2014,dunn_2017}, the process of construction learning \cite{Barak2017, Barak2017a}, and whether constructional information is implicitly encoded in models like BERT \cite{tayyar-madabushi-etal-2020-cxgbert}.

A commonly discussed example of a construction is the \textit{ditransitive} in (a1). CxGs use a constraint-based formalism in which each slot in the construction is defined by a particular slot-constraint; in (a1) these are syntactic constraints. One of the important ideas in CxG is that constructions themselves carry a meaning. For example, the ditransitive construction carries a meaning of \textsc{transfer} regardless of the meaning of the particular verb that is used in the ditransitive. In some cases, this notion of transfer also follows from the meaning of the verb, as in (a2) with \textit{sold}. But, in other cases, utterances like (a3) can take on a meaning of transfer that is not present in the verb \textit{smile}. Constructions can also have idiomatic children, such as (a4), in which an item-specific slot-constraint defines a sub-class of the construction, like (a5), whose meaning is not entirely predictable or transparent given the parent construction.

~

\noindent (a1) [ Syn: \textsc{np} -- Syn: \textsc{vp} -- Syn: \textsc{np} -- Syn: \textsc{np} ]

\noindent (a2) ``He sold them a car."

\noindent (a3) ``He smiled himself an upgrade."

\noindent (a4) ``He gave them a hand."

\noindent (a5) [ Syn: \textsc{np} -- Lex: \textit{give} -- Syn: \textsc{np} -- Lex: \textit{hand} ]

~

\begin{table}[t]
\centering
\begin{tabular}{|c|c|c|}
\hline
\textbf{Language} & \textbf{Code} & \textbf{Family} \\
\hline
Danish & dan & Germanic \\
Dutch & nld & Germanic \\
English & eng & Germanic \\
German & deu & Germanic \\
Norwegian & nor& Germanic \\
Swedish & swe & Germanic \\
\hline
Catalan & cat & Romance \\
French & fra & Romance \\
Italian & ita & Romance \\
Portuguese & por & Romance \\
Romanian & ron & Romance \\
Spanish & spa & Romance \\
\hline
  \end{tabular}
  \caption{Sources of Language Data, with 2 million words each for the \textsc{tw, wk}, and \textsc{cc} registers}
  \label{tab:1}
\end{table}

Register is a distinct pattern of usage that is associated with the context of production. A substantial body of research has shown that register is a major source of linguistic variation \citep{Biber2012}. Recent work has shown that the impact of register variation exceeds the impact of geographic variation in many cases \citep{Dunn2021}. The result of register variation is that large corpora often contain a number of distinct sub-corpora, each with their own unique patterns of usage \citep{Sardinha2018, Cvrcek2020}. In other words, a gigaword web-crawled corpus is not simply a flat collection of many written documents: there is, instead, a register-based grouping of sub-corpora which often contain significantly different linguistic forms.

This paper simulates the acquisition of constructions by incrementally increasing the amount of exposure: 100k words, 200k words, 300k words and so on up to 2 million words \cite{Alishahi2008, Matusevych2013, Beekhuizen2015}. This provides a series of grammars, each representing a different state in the learning process. This experiment is repeated across three registers, each with a unique set of constructional frequencies: Wikipedia (formal), Twitter (informal), and Web (mixed). Each register has a progression of 20 register-specific grammars, with each grammar representing different levels of exposure.

Is there a level of exposure at which register-specific grammars reach a stable shared representation of the language? On the one hand, it is possible that register-specific grammars are maintained as unique sub-sets of linguistic behaviour. In this case, grammars \textit{would not} converge given increased exposure. On the other had, it is possible that register-specific constructions fade away as increased exposure leads to more generalized grammars. In this case, grammars \textit{would} converge, becoming more similar and less register-specific as they are learned through exposure to more training data.

To avoid language-specific generalizations, this experiment is repeated across six Germanic languages (Danish, Dutch, English, German, Norwegian, Swedish) and six Romance languages (Catalan, French, Italian, Portuguese, Romanian, Spanish). Each grammar contains a set of constructions that have been learned to best represent the training data. Thus, for each stage in the learning process, we can measure the overlap between register-specific grammars (Twitter to Web, Twitter to Wikipedia, and Wikipedia to Web). When register-specific grammars have a \textit{higher overlap} this means that they share more of their constructional representations. In other words, higher overlap means that the grammars are more similar.

We say that grammars \textit{converge} when they have a higher degree of overlap or similarity. This paper develops two measures of grammar similarity to capture different aspects of convergence: a fuzzy Jaccard similarity that captures convergence across even rare constructions and a frequency-weighted Jaccard similarity that focuses on the core constructions. These two measures of convergence allow us to model the degree to which construction grammars learn register-specific representations.

\begin{table*}[t]
\centering
\begin{tabular}{|cl|cl|}
\hline
~ & \textbf{Construction (Type)} & ~ & \textbf{Construction (Type)} \\
\hline
(b) & [ Syn: \textsc{n} - Lex: \textit{of} - Syn: \textsc{det} - Sem:<587> ] & (c) & [ Lex: \textit{while} - Sem:<113> - Syn: \textsc{adp} ] \\
\hline
~ & \textbf{Examples (Tokens)} & ~ & \textbf{Examples (Tokens)} \\
\hline
(b1) & `spirit of the alchemist' & (c1) & `while working out' \\
(b2) & `provinces of the empire' & (c2) & `while sitting by' \\
(b3) & `raft of the medusa' & (c3) & `while going through' \\
(b4) & `constellations of the zodiac' & (c4) & `while carrying out' \\
(b5) & `myth of the anaconda' & (c5) & `while sticking around' \\
\hline
  \end{tabular}
  \caption{Examples of Constructions (Types) and Instances of Constructions (Tokens) for English}
  \label{tab:2}
\end{table*}

There are three possible outcomes for this experimental framework: \textit{First}, it is possible that grammars converge as exposure increases. This convergence would indicate that register variation becomes less important given more data: the grammars contain the same constructions regardless of the input register. In other words, this would indicate that more data overall is able to compensate for register variation. \textit{Second}, it is possible that grammars do not converge as the amount of exposure increases. This outcome would indicate that each register represents a unique sub-grammar, a distinct set of linguistic behaviours. In this case, more data overall would never compensate for register variation. These two competing hypotheses are tested in Experiment 1 (Section 5). This experiment  finds that increased exposure does, in fact, lead to converging grammars. This finding supports the first hypothesis that is described above.

\textit{Third}, another possible outcome is that not all languages pattern together. In other words, it may be the case that some Germanic languages converge given increased exposure but some Romance languages retain register-specific constructions. Part of the experimental design is to repeat the same framework across many languages to determine whether the outcome is specific to one or another language. We will see, in Experiment 1, that there is variation across languages in both (i) the rate of convergence and (ii) the upper limit when grammars stop converging. However, it remains the case that all languages show increased convergence given increased exposure.

Given that the first experiments show that grammars do converge given increased exposure, we undertake an additional experiment: is there a core construction grammar for each language? Construction grammars have a proto-type structure, meaning that some representations are central (and thus very frequent) while others are peripheral (and thus somewhat rare). We use the frequency-weighted Jaccard similarity in Experiment 2 (Section 6) to determine whether the overall rate of convergence changes when we focus on the core of the grammar rather than the periphery. These experiments show that for most languages the core grammar is acquired very early with little change in convergence given increased exposure.

\section{Experimental Design}

The basic experimental approach is to learn grammars over increasing amounts of exposure: from 100k words to 2 million words in increments of 100k words (thus creating 20 grammars per condition). This series of grammars simulates the accumulation of grammatical knowledge as the amount of exposure increases. This approach is repeated across each of the three registers that represent different contexts of production.

The register-specific data used to progressively learn grammars is collected from three sets of corpora: tweets (\textsc{tw}), Wikipedia articles (\textsc{wk}), and web pages (\textsc{cc} for Common Crawl). This dataset is summarized in Table \ref{tab:1}. The corpus contains the same amount of data per register per language \cite{Dunn2020, dunn-adams-2020-geographically}.

The pairwise similarity relationships between grammars differ, in part, because some sources of data are more similar to one another. For example, \textsc{cc} and \textsc{wk} are more similar registers and thus their grammars have a baseline similarity that is higher than \textsc{wk} and \textsc{tw}. In other words, the Wikipedia grammars (CxG-\textsc{wk}) are more similar to the web grammars (CxG-\textsc{cc}) than to the Twitter grammars (CxG-\textsc{tw}). What matters, then, is the degree to which the \textit{relative} similarity between grammars changes as the amount of exposure increases. This approach controls for the underlying similarity between registers.

\section{Learning Constructions}
\label{section:learning-constructions}

The grammar induction algorithm used to learn constructions is taken from previous work ~\cite{dunn_2017}. At its core, this model of CxGs has three main components: \textit{First}, a psychologically-plausible measure of association, the $\Delta P$, is used to measure the entrenchment of particular representations \cite{Ellis2007,Gries2013,DunnIJCL}. \textit{Second}, an association-based beam search is used to identify constructions of arbitrary length by finding the most entrenched representation for each training sentence in reference to a matrix of $\Delta P$ values \cite{Dunn2019a}. \textit{Third}, a Minimum Description Length measure is used as an optimization function, balancing the trade-off between increased storage of item-specific constructions and increased computation of generalized constructions \cite{d18}. We briefly review each of these components of the algorithm in this section.

Constructions are constraint-based syntactic representations in which individual slots are limited to particular syntactic, semantic, or lexical items ~\cite{goldberg1995constructions}. Construction Grammars are \textit{usage-based} in the sense that constructions range from very idiomatic phrases (like ``give me a hand") to very abstract sequences (like \textsc{np -> det adj np}). One of the many advantages of CxGs is that they represent actual usage, which means that they are capable of identifying syntactic variation across dialects \cite{d18b,Dunn2019ab,10.3389/frai.2019.00015} and even across individuals \cite{Nini2021}. But the disadvantage is that the induction algorithm must learn both the units of representation (i.e., semantic categories) as well as these multi-dimensional slot-constraints. For example, the constructions in (b) and (c) in Table \ref{tab:2} include all three types of representation (lexical, syntactic, semantic) so that the algorithm must be able to navigate across these three representations during the learning process.

The grammar induction algorithm starts by defining the types of representation. \textsc{lexical} constraints are based on word-forms, without lemmatization. These are the simplest and most idiomatic types of constraints. \textsc{syntactic} constraints are formulating using the universal part-of-speech tagset \cite{pdm12} and implemented using the Ripple Down Rules algorithm \cite{nn}. \textsc{semantic} constraints are based on distributional semantics, with k-means clustering applied to discretize pre-trained fastText embeddings \cite{Grave2019}. The semantic constraints in (b) and (c) are formulated using the index of the corresponding clusters. A complete list of semantic domains used in this paper, along with a grammar for English, are available in the supplementary material.

Each sentence in the input corpus is transformed into these three parallel dimensions of representation (lexical, syntactic, semantic). In the first stage of the algorithm, a co-occurrence matrix is produced that represents the association between all pairs of representations using the $\Delta P$ measure, shown below. What distinguishes the $\Delta P$ from more common measures like PPMI~\cite{church-hanks-1989-word,dagan-etal-1993-contextual} is that it has direction-specific variants that take ordering into account, thus helping to capture syntactic patterns. The measure, calculated left-to-right, is the probability that two units occur together ($X$ and $Y$) adjusted by the probability that $X$ occurs alone. In this notation, $Y_P$ indicates that unit $Y$ is present and $Y_A$ that unit $Y$ is absent.

\begin{equation}
\Delta P_{LR} = p(X_P|Y_P)-p(X_P|Y_A)
\end{equation}

Given (i) a three-dimensional representation of each sentence and (ii) a co-occurrence matrix with the directional association for each pair of representations, a beam search algorithm is used to find the most \textit{entrenched} sequence of constraints for each sentence in the training corpus. The basic idea behind this search is to traverse all possible paths of constraints, ending each path when the cumulative $\Delta P$ falls below a threshold \cite{Dunn2019a}. For each sentence, the sequence of slot-constraints with the highest cumulative association is added to a provisional grammar. In CxG, some representations are very entrenched (grammaticalized) and others are only slightly entrenched \cite{Goldberg2011,Goldberg2016}. The optimum sub-set of constructions is then selected from this provisional grammar.

The grammar induction model itself is based on the Minimum Description Length paradigm \cite{GrunwaldPandRissanen2007, Goldsmith2001, Goldsmith2006}. In this kind of model, observed probabilities are used to calculate the encoding size of a grammar ($L_1$) as well as the encoding size of a test corpus given that grammar ($L_2$). Usage-based grammar posits a trade-off between memory and computation; this is modelled by MDL's combination of $L_1$ and $L_2$ encoding size \cite{d18}.

The best grammar is the one which minimizes this metric on a test corpus. In (2), $G$ refers to the grammar being evaluated and $D$ refers to the test corpus. For example, this is used to choose the parameters of the beam search algorithm described above. In practice, the use of MDL to evaluate grammars is quite similar to the use of perplexity to evaluate language models; for example, the MDL metric is specific to each test corpus. The advantage of the MDL metric for usage-based grammar is that it distinguishes between the complexity of the grammar ($L_1$) and the fit between the grammar and the test corpus ($L_2$).

\begin{equation}
MDL = \min\limits_{G} \{{L_1(G)+L_2 (D \mid G)}\}
\end{equation}

This induction algorithm provides a grammar of constructions to describe the training corpus, where the grammar is chosen to minimize the MDL metric. The grammar is a set of constructions, each of which is a sequence of slot-constraints. And each slot-constraint is formulated using the basic inventory of lexical, syntactic, and semantic fillers.

In previous work, the induction algorithm used alternating training and testing sets to refine grammars. A large background corpus was used to estimate the $\Delta P$ matrix that guides the selection of slot-constraints. The experiments here, however, depend on limiting the amount of training data as a means of controlling for different levels of exposure. In each condition, then, the same training data is used for each stage in the algorithm. In other words, the 200k word exposure condition has access only to the 200k word training corpus (with the implicit exception of the pre-trained embeddings). Each model is trained on the same underlying corpus, so that the 500k word condition is given the same data as the 400k word condition plus an additional 100k words of new data.

\begin{figure*}[t]
\centering
\includegraphics[width = 450pt]{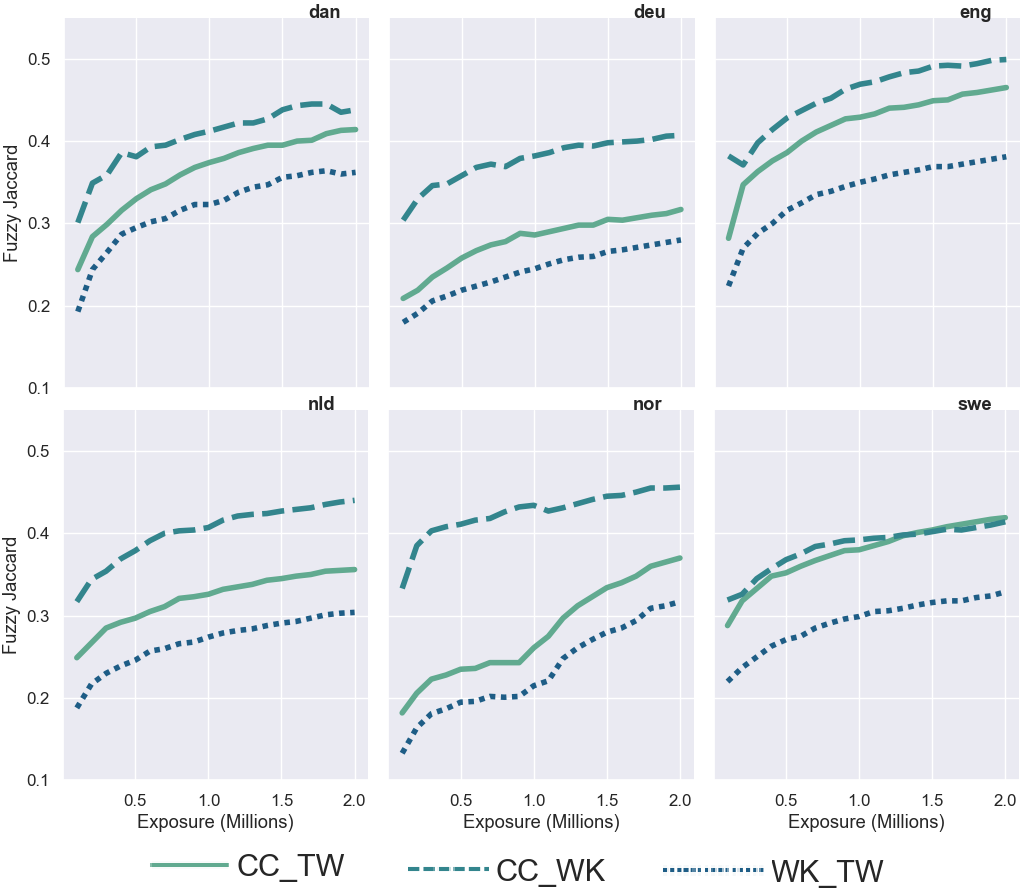}
\caption{Overlap between pairs of register-specific grammars by amount of exposure, using Fuzzy Jaccard Similarity for Germanic languages to measure Constructional Overlap (Experiment 1). Each line represents the similarity between two grammars learned from different registers.}
\label{fig:fuzzy_germanic}
\end{figure*}

\begin{figure*}[t]
\centering
\includegraphics[width = 450pt]{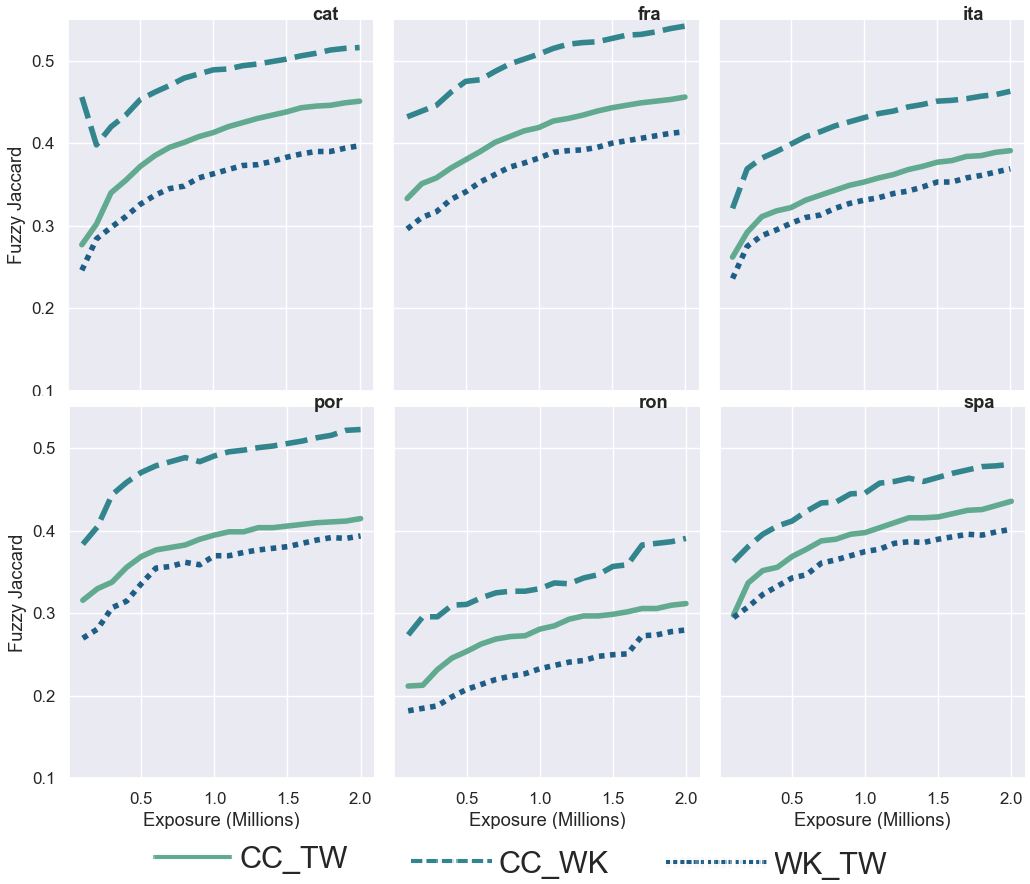}
\caption{Overlap between pairs of register-specific grammars by amount of exposure, using Fuzzy Jaccard Similarity for Romance languages to measure Constructional Overlap (Experiment 1). Each line represents the similarity between two grammars learned from different registers.}
\label{fig:fuzzy_romance}
\end{figure*}

\section{Measuring Grammar Similarity}

A grammar in this context is a set of constructions, where each construction is a sequence of slot-constraints. Our central measure of overlap between grammars, then, is based on the Jaccard similarity, where values close to 1 represent very similar grammars and values close to 0 represent very different grammars.

\begin{equation}
   J(A, B) = \frac{|A\cap B|}{|A \cup B|}
\end{equation}

One of the challenges with CxGs is that two different representations could use different slot-constraints to capture a similar set of utterances, essentially providing two versions of the same construction. Consider the construction in (d), a part of the English web-based grammar. The tokens in (d1) through (d4) are tokens of this construction. An alternate constraint that specifies \textit{the} rather than \textsc{det} might be chosen, leading to a different representation for the same underlying construction.

~

\noindent (d) [ Lex: \textit{how to} -- Syn: \textsc{v} -- Syn: \textsc{det} -- Syn: \textsc{n} ]

\noindent (d1) `how to get the job'

\noindent (d2) `how to track a phone'

\noindent (d3) `how to improve the system'

\noindent (d4) `how to start a blog'

~

The challenge for calculating convergence using the Jaccard similarity between grammars is that similar constructions could capture the same set of tokens. For example, the syntactic constraint (\textsc{det}) could be replaced with a lexical constraint (\textit{the}) in the construction in (d). The Jaccard similarity on its own would not capture the similarity between these two alternate formulations of what is ultimately the same underlying construction.

Our first measure is thus a \textsc{fuzzy jaccard similarity}, in which the definition of set membership is extended to very similar constructions. In this measure, a sub-sequence matching algorithm is used to find how many slot-constraints are shared between two constructions, taking order into account. Any two constructions above a threshold of 0.71 shared sub-sequences are considered a match. This threshold is chosen because it allows one slot-constraint to differ between most constructions while still considering them to be similar representations. For example, two six-slot constructions must share five constraints in order to count as a match at this threshold. This measure thus provides a better approximation of construction similarity, focusing on constructions with slightly different internal constraints or with an added slot-constraint in one position.

Our second measure is a \textsc{frequency-weighted jaccard similarity}, in which the importance of each construction is weighted relative to its frequency in an independent corpus. For each language, a background corpus is created from a mix of different registers: Open Subtitles and Global Voices (news articles) \cite{Tiedemann2012} and Bible translations \cite{ChristodoulopoulosC.andSteedman2015}. This background corpus represents usage external to the three main registers used in the experiments. 

The frequency of each construction is derived from 500k words of this background corpus, so that very common constructions are given more weight in the similarity measure. The basic idea is that some constructions are part of the core grammar, thus being frequent in all registers. This weighted measure captures convergence within this core grammar by focusing on those constructions which are most frequent in an independent corpus.

These two measures based on Jaccard similarity provide three values for each condition: \textsc{cc-wk}, \textsc{cc-tw}, and \textsc{wk-tw}. Each of the values represents a pairwise similarity between two register-specific grammars. The higher these values, the more the learner is converging onto a shared grammar.

\begin{figure*}[t]
\centering
\includegraphics[width = 450pt]{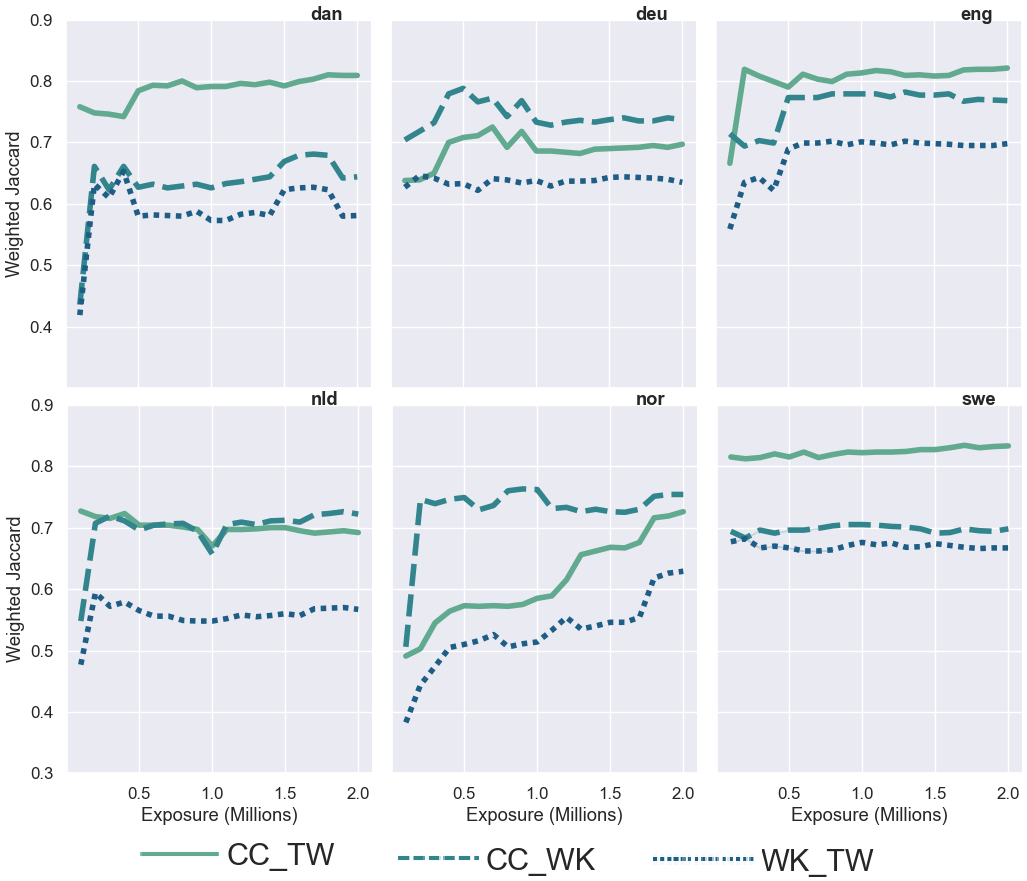}
\caption{Overlap between pairs of register-specific grammars by amount of exposure, using Weighted Jaccard Similarity for Germanic languages to measure Core Grammatical Overlap (Experiment 2). Each line represents the similarity between two grammars learned from different registers.}
\label{fig:weighted_germanic}
\end{figure*}

\begin{figure*}[t]
\centering
\includegraphics[width = 450pt]{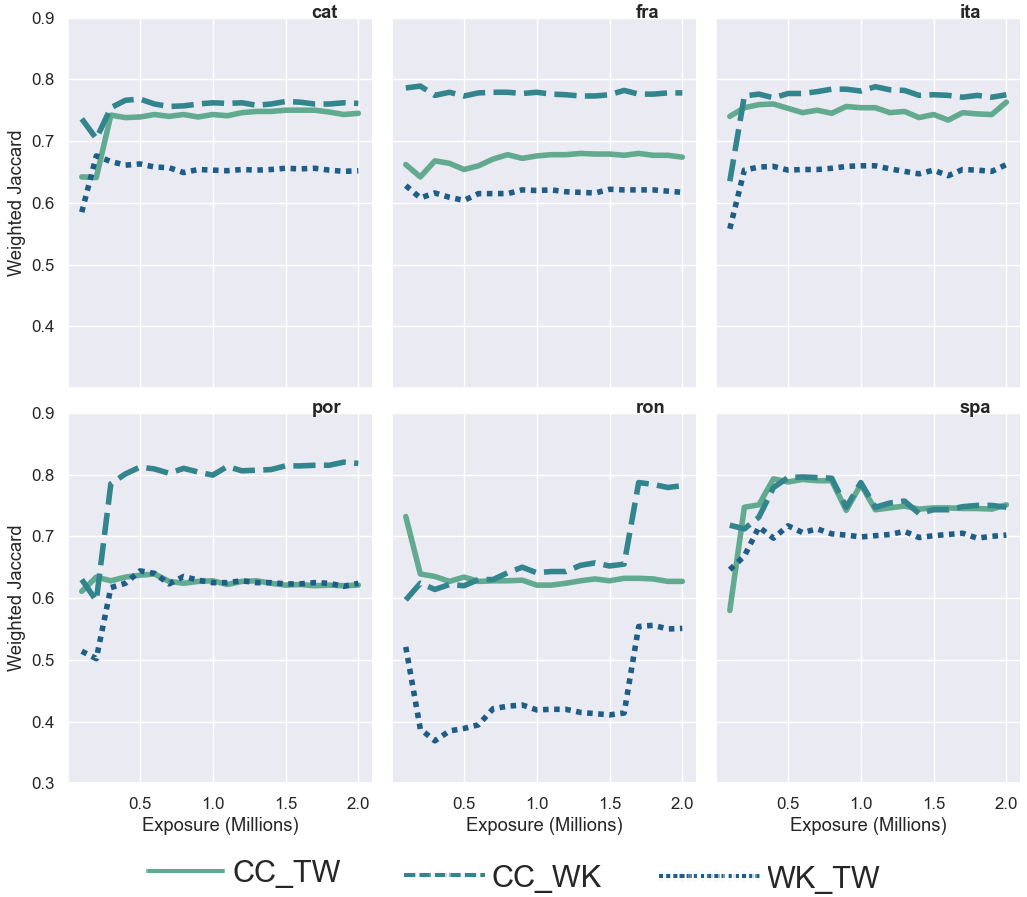}
\caption{Overlap between pairs of register-specific grammars by amount of exposure, using Weighted Jaccard Similarity for Romance languages to measure Core Grammatical Overlap (Experiment 2). Each line represents the similarity between two grammars learned from different registers.}
\label{fig:weighted_romance}
\end{figure*}

We have thus formulated two measures of grammatical overlap. The first, fuzzy Jaccard, captures the overall similarity between grammars. The second, frequency-weighted Jaccard, captures the similarity between grammars with a focus on the core constructions that are frequent across registers (ignoring the long tail of register-specific forms). The following sections apply these measures of grammatical overlap to the CxGs exposed to increasing amounts of usage. In each case, higher values represent increased convergence between grammars.

\section{Experiment 1: Constructional Overlap}

To what degree do grammars converge onto shared representations as they are exposed to increasing amounts of data from different registers? The first experiment uses the fuzzy Jaccard similarity to model convergence. The results are shown in Figure \ref{fig:fuzzy_germanic} and Figure \ref{fig:fuzzy_romance}. In both cases, the y axis represents the similarity between register-specific grammars, with higher values representing more convergent sets of constructions. Note that each line represents the similarity between two register-specific grammars. And the x axis represents the amount of exposure, moving from 100k words up to 2 million words. Languages are labelled using their language code, as listed in Table \ref{tab:1}.

We notice, first, that there is a baseline difference in the similarity between registers. In every language, for example, \textsc{wk} and \textsc{tw} are the least similar. And, in every language, \textsc{cc} and \textsc{wk} are the most similar. This pattern is shared across all 12 languages because the underlying contexts of production have this similarity, with Wikipedia the most formal and Twitter the least formal. The distance between registers does not matter here. What does matter is the relative change in distance as the algorithm is exposed to more data (i.e., \textsc{cc-tw} at 200k words and \textsc{cc-tw} at 2 million words).

We notice, second, that in every language grammars converge with increased exposure. The overall similarity between constructions increases as more training data is observed. This is true for both Germanic and Romance languages, showing this to be a robust generalization.

Languages do differ, however, in (i) the overall amount of similarity and (ii) the rate of convergence. In the first case, within Germanic languages the range of grammar similarity is generally comparable, starting at approximately 0.2 and ending at approximately 0.4. Some languages (like English) reach a higher level of convergence. Other languages have register-specific patterns: in Norwegian the similarity between \textsc{cc-wk} is quite high throughout, but in Swedish the similarity between \textsc{cc-wk} is the same as the similarity between \textsc{cc-tw}. In other words, the ordering of register similarity is constant across languages but the distance is not. Romance languages have a generally higher overall rate of similarity, although there is a wide gap between French (the highest) and Romanian (the lowest). This means that there is some variation across languages in terms of how similar the register-specific grammars become.

There is also variation in the rate of convergence. Some languages (like Swedish) have a somewhat flat rate of convergence while other languages (like English) have a steeper curve. A flat curve represents a slow convergence while a steep curve represents a rapid convergence as exposure increases. Some languages have bursts of convergence at specific amounts of exposure. For example, Norwegian has a steep increase until about 300k words, then remains flat until about 1 million words before beginning to converge again. In other words, if we think about the growth curve of similarity, the pattern of convergence differs across languages. The cause of these differences is a matter for future research. The basic conclusion here is that all languages show increasing convergence with increasing exposure. In other words, register-specific syntactic patterns generalize as the induction algorithm encounters more training data.

\section{Experiment 2: The Core Constructions}

Although many constructions have varying frequency across different registers, we would expect that the grammar of each language also has a core set of very frequent constructions which are shared across all registers. In other words, register variation itself should not be strong enough to erase the syntactic generalizations provided by a grammar. The frequency-weighted Jaccard similarity measure is used to find this core set of constructions: how much do grammars change with increasing exposure when we focus on the most frequent constructions? As explained above, the frequency weighting is derived from independent corpora that represent a different set of registers.

The weighted similarity measures are shown in Figure \ref{fig:weighted_germanic} and Figure \ref{fig:weighted_romance}, again with the y axis showing similarity (higher values mean more overlap between grammars) and the x axis showing increasing amounts of exposure. Each line represents the similarity between two register-specific grammars. The overall level of similarity is significantly higher here. The Germanic languages range from approximately 0.7 (Dutch) to approximately 0.8 (Swedish). The Romance languages have the same range, with Spanish the lowest and Portuguese the highest. This overall increase in similarity shows that, when focusing on the core constructions, the register-specific grammars converge quickly.

We notice, second, that the growth curve for the frequency-weighted measure shows very little change after a certain point. For many languages, like Portuguese and Italian, there is a sharp increase after several hundred thousand words of exposure. This indicates that the initial grammars (based on low exposure) are not adequate. After more exposure, however, the stable core of constructions is acquired. Once that initial burst of acquisition is complete, there are no significant changes. This is not true for Norwegian (which continues to show a continuous growth of convergence) or for Romanian (which has an initial decline and a much later burst of similarity). But the overall pattern across all languages is that the core set of constructions remains stable after a small amount of exposure.

\section{Conclusions}

These experiments show that register-specific grammars converge onto shared constructions as they are exposed to more training data. This is observed across 12 languages and three registers. At the same time, each language has a core set of frequent constructions which is \textit{not} influenced by register variation. This core CxG is acquired for each language given a limited amount of exposure and does not change significantly as exposure increases. These results are important for describing the interaction between syntactic generalizations (the core grammars) and syntactic variation (the register-specific grammars).

\bibliography{anthology,custom}

\begin{thebibliography}{43}
\expandafter\ifx\csname natexlab\endcsname\relax\def\natexlab#1{#1}\fi

\bibitem[{Alishahi and Stevenson(2008)}]{Alishahi2008}
A.~Alishahi and S.~Stevenson. 2008.
\newblock \href
  {https://citeseerx.ist.psu.edu/viewdoc/summary?doi=10.1.1.126.418} {{A
  Computational Model of Early Argument Structure Acquisition}}.
\newblock \emph{Cognitive Science}, 32(5):789--834.

\bibitem[{Barak and Goldberg(2017)}]{Barak2017}
L.~Barak and A~Goldberg. 2017.
\newblock \href
  {https://www.aaai.org/ocs/index.php/SSS/SSS17/paper/download/15297/14526}
  {{Modeling the Partial Productivity of Constructions}}.
\newblock In \emph{Proceedings of AAAI 2017 Spring Symposium on Computational
  Construction Grammar and Natural Language Understanding}, pages 131--138.
  Association for the Advancement of Artificial Intelligence.

\bibitem[{Barak et~al.(2017)Barak, Goldberg, and Stevenson}]{Barak2017a}
L.~Barak, A.~Goldberg, and S.~Stevenson. 2017.
\newblock \href {http://dx.doi.org/10.18653/v1/D16-1010} {{Comparing
  Computational Cognitive Models of Generalization in a Language Acquisition
  Task}}.
\newblock In \emph{Proceedings of the Conference on Empirical Methods in NLP},
  pages 96--106. Association for Computational Linguistics.

\bibitem[{Beckner et~al.(2009)Beckner, Ellis, Blythe, Holland, Bybee, Ke,
  Christiansen, Larsen-Freeman, Croft, and Schoenemann}]{Beckner2009}
C.~Beckner, N.~Ellis, R.~Blythe, J.~Holland, J.~Bybee, J.~Ke, M.~Christiansen,
  D.~Larsen-Freeman, W.~Croft, and T.~Schoenemann. 2009.
\newblock \href
  {http://www.unm.edu/{~}jbybee/downloads/BecknerEtAl2009ComplexAdaptiveSystem.pdf}
  {{Language Is a Complex Adaptive System: Position Paper}}.
\newblock \emph{Language Learning}, 59:1--26.

\bibitem[{Beekhuizen et~al.(2015)Beekhuizen, Bod, Fazly, Stevenson, and
  Verhagen}]{Beekhuizen2015}
B.~Beekhuizen, R.~Bod, A.~Fazly, S.~Stevenson, and A.~Verhagen. 2015.
\newblock \href {https://doi.org/10.3115/v1/w14-2006} {{A Usage-Based Model of
  Early Grammatical Development}}.
\newblock In \emph{Proceedings of the Workshop on Cognitive Modeling and
  Computational Linguistics}, pages 46--54. Association for Computational
  Linguistics.

\bibitem[{Biber(2012)}]{Biber2012}
D.~Biber. 2012.
\newblock \href {https://doi.org/10.1515/cllt-2012-0002} {{Register as a
  predictor of linguistic variation.}}
\newblock \emph{Corpus Linguistics and Linguistic Theory}, 8(1):9--37.

\bibitem[{Bybee(2006)}]{Bybee2006}
J.~Bybee. 2006.
\newblock \href {https://www.jstor.org/stable/4490266} {{From Usage to Grammar:
  The mind's response to repetition}}.
\newblock \emph{Language}, 82(4):711--733.

\bibitem[{Christodoulopoulos and
  Steedman(2015)}]{ChristodoulopoulosC.andSteedman2015}
C.~Christodoulopoulos and M.~Steedman. 2015.
\newblock \href
  {https://link.springer.com/content/pdf/10.1007/s10579-014-9287-y.pdf} {{A
  massively parallel corpus: The Bible in 100 languages}}.
\newblock \emph{Language Resources and Evaluation}, 49:375--395.

\bibitem[{Church and Hanks(1989)}]{church-hanks-1989-word}
K.~Church and P.~Hanks. 1989.
\newblock \href {https://doi.org/10.3115/981623.981633} {Word association
  norms, mutual information, and lexicography}.
\newblock In \emph{27th Annual Meeting of the Association for Computational
  Linguistics}, pages 76--83. Association for Computational Linguistics.

\bibitem[{Cvr{\v{c}}ek et~al.(2020)Cvr{\v{c}}ek, Komrskov{\'{a}}, Luke{\v{s}},
  Poukarov{\'{a}}, Řehořkov{\'{a}}, Zasina, and Benko}]{Cvrcek2020}
V.~Cvr{\v{c}}ek, Z.~Komrskov{\'{a}}, D.~Luke{\v{s}}, P.~Poukarov{\'{a}},
  A.~Řehořkov{\'{a}}, A.~Zasina, and V.~Benko. 2020.
\newblock \href {https://doi.org/10.1007/s10579-020-09487-4} {{Comparing
  web-crawled and traditional corpora}}.
\newblock \emph{Language Resources and Evaluation}, 54(3):713--745.

\bibitem[{Dagan et~al.(1993)Dagan, Marcus, and
  Markovitch}]{dagan-etal-1993-contextual}
I.~Dagan, S.~Marcus, and S.~Markovitch. 1993.
\newblock \href {https://doi.org/10.3115/981574.981596} {Contextual word
  similarity and estimation from sparse data}.
\newblock In \emph{31st Annual Meeting of the Association for Computational
  Linguistics}, pages 164--171. Association for Computational Linguistics.

\bibitem[{Dunn(2017)}]{dunn_2017}
J.~Dunn. 2017.
\newblock \href {https://doi.org/10.1017/langcog.2016.7} {Computational
  learning of construction grammars}.
\newblock \emph{Language \& Cognition}, 9(2):254–292.

\bibitem[{Dunn(2018{\natexlab{a}})}]{d18b}
J.~Dunn. 2018{\natexlab{a}}.
\newblock \href {https://doi.org/10.1515/cog-2017-0029} {{Finding Variants for
  Construction-Based Dialectometry: A Corpus-Based Approach to Regional CxGs}}.
\newblock \emph{Cognitive Linguistics}, 29(2):275--311.

\bibitem[{Dunn(2018{\natexlab{b}})}]{d18}
J.~Dunn. 2018{\natexlab{b}}.
\newblock \href {https://www.aclweb.org/anthology/W18-0309} {{Modeling the
  Complexity and Descriptive Adequacy of Construction Grammars}}.
\newblock In \emph{Proceedings of the Society for Computation in Linguistics},
  pages 81--90. Association for Computational Linguistics.

\bibitem[{Dunn(2018{\natexlab{c}})}]{DunnIJCL}
J.~Dunn. 2018{\natexlab{c}}.
\newblock \href {https://doi.org/10.1075/ijcl.16098.dun} {Multi-unit
  association measures: Moving beyond pairs of words}.
\newblock \emph{International Journal of Corpus Linguistics}, 23:183--215.

\bibitem[{Dunn(2019{\natexlab{a}})}]{Dunn2019a}
J.~Dunn. 2019{\natexlab{a}}.
\newblock \href {https://www.aclweb.org/anthology/W19-2913.pdf} {{Frequency vs.
  Association for Constraint Selection in Usage-Based Construction Grammar}}.
\newblock In \emph{Proceedings of the Workshop on Cognitive Modeling and
  Computational Linguistics}. Association for Computational Linguistics.

\bibitem[{Dunn(2019{\natexlab{b}})}]{10.3389/frai.2019.00015}
J.~Dunn. 2019{\natexlab{b}}.
\newblock \href {https://doi.org/10.3389/frai.2019.00015} {{Global Syntactic
  Variation in Seven Languages: Toward a Computational Dialectology}}.
\newblock \emph{Frontiers in Artificial Intelligence}, 2:15.

\bibitem[{Dunn(2019{\natexlab{c}})}]{Dunn2019ab}
J.~Dunn. 2019{\natexlab{c}}.
\newblock \href {http://dx.doi.org/10.18653/v1/W19-1405} {{Modeling Global
  Syntactic Variation in English Using Dialect Classification}}.
\newblock In \emph{Proceedings of Workshop on NLP for Similar Languages,
  Varieties and Dialects}, pages 42--53. Association for Computational
  Linguistics.

\bibitem[{Dunn(2020)}]{Dunn2020}
J.~Dunn. 2020.
\newblock \href {https://doi.org/10.1007/s10579-020-09489-2} {{Mapping
  languages: The Corpus of Global Language Use}}.
\newblock \emph{Language Resources and Evaluation}, 54:999–1018.

\bibitem[{Dunn(2021)}]{Dunn2021}
J.~Dunn. 2021.
\newblock \href {https://aclanthology.org/2021.vardial-1.4} {Representations of
  language varieties are reliable given corpus similarity measures}.
\newblock In \emph{Proceedings of the Eighth Workshop on NLP for Similar
  Languages, Varieties and Dialects}, pages 28--38. Association for
  Computational Linguistics.

\bibitem[{Dunn and Adams(2020)}]{dunn-adams-2020-geographically}
J.~Dunn and B.~Adams. 2020.
\newblock \href {https://www.aclweb.org/anthology/2020.lrec-1.308}
  {Geographically-balanced gigaword corpora for 50 language varieties}.
\newblock In \emph{Proceedings of the 12th Language Resources and Evaluation
  Conference}, pages 2528--2536. European Language Resources Association.

\bibitem[{Dunn and Nini(2021)}]{Nini2021}
J.~Dunn and A.~Nini. 2021.
\newblock \href {https://www.aclweb.org/anthology/2021.cmcl-1.19.pdf}
  {{Production vs Perception: The Role of Individuality in Usage-Based Grammar
  Induction}}.
\newblock In \emph{Proceedings of the Workshop on Cognitive Modeling and
  Computational Linguistics}, pages 149--159. Association for Computational
  Linguistics.

\bibitem[{Ellis(2007)}]{Ellis2007}
N.~Ellis. 2007.
\newblock \href {https://doi.org/10.1093/applin/ami038} {{Language Acquisition
  as Rational Contingency Learning}}.
\newblock \emph{Applied Linguistics}, 27(1):1--24.

\bibitem[{Fodor and Crowther(2002)}]{fodor2002understanding}
J.~Fodor and C.~Crowther. 2002.
\newblock \href
  {https://citeseerx.ist.psu.edu/viewdoc/summary?doi=10.1.1.424.5766}
  {{Understanding Stimulus Poverty Arguments}}.
\newblock \emph{The Linguistic Review}, 19(1-2):105--145.

\bibitem[{Forsberg et~al.(2014)Forsberg, Johansson, Bckstrm, Borin, Lyngfelt,
  Olofsson, and Prentice}]{Forsberg2014}
M.~Forsberg, R.~Johansson, L.~Bckstrm, L.~Borin, B.~Lyngfelt, J.~Olofsson, and
  J.~Prentice. 2014.
\newblock \href {https://doi.org/10.1075/cf.6.1.07for} {{From Construction
  Candidates to Constructicon Entries: An experiment using semi-automatic
  methods for identifying constructions in corpora}}.
\newblock \emph{Constructions and Frames}, 6(1):114--135.

\bibitem[{Goldberg(1995)}]{goldberg1995constructions}
A.~Goldberg. 1995.
\newblock \href {https://books.google.co.uk/books?id=HzmGM0qCKtIC}
  {\emph{Constructions: A Construction Grammar Approach to Argument
  Structure}}.
\newblock Cognitive Theory of Language and Culture Series. University of
  Chicago Press.

\bibitem[{Goldberg(2006)}]{g06a}
A.~Goldberg. 2006.
\newblock \emph{{Constructions at work: The nature of generalization in
  language}}.
\newblock Oxford University Press, Oxford.

\bibitem[{Goldberg(2011)}]{Goldberg2011}
A.~Goldberg. 2011.
\newblock \href {https://doi.org/10.1515/cogl.2011.006} {{Corpus evidence of
  the viability of statistical preemption}}.
\newblock \emph{Cognitive Linguistics}, 22(1):131--154.

\bibitem[{Goldberg(2016)}]{Goldberg2016}
A.~Goldberg. 2016.
\newblock \href {https://doi.org/10.1017/langcog.2016.17} {{Partial
  Productivity of Linguistic Constructions: Dynamic categorization and
  Statistical preemption}}.
\newblock \emph{Language {\&} Cognition}, 8(3):369--390.

\bibitem[{Goldsmith(2001)}]{Goldsmith2001}
J.~Goldsmith. 2001.
\newblock \href {https://doi.org/10.1162/089120101750300490} {{Unsupervised
  Learning of the Morphology of a Natural Language}}.
\newblock \emph{Computational Linguistics}, 27(2):153--198.

\bibitem[{Goldsmith(2006)}]{Goldsmith2006}
J.~Goldsmith. 2006.
\newblock \href
  {https://citeseerx.ist.psu.edu/viewdoc/download?doi=10.1.1.127.8428&rep=rep1&type=pdf}
  {{An Algorithm for the Unsupervised Learning of Morphology}}.
\newblock \emph{Natural Language Engineering}, 12(4):353--371.

\bibitem[{Grave et~al.(2019)Grave, Bojanowski, Gupta, Joulin, and
  Mikolov}]{Grave2019}
E.~Grave, P.~Bojanowski, P.~Gupta, A.~Joulin, and T.~Mikolov. 2019.
\newblock \href {http://arxiv.org/abs/1802.06893} {{Learning word vectors for
  157 languages}}.
\newblock In \emph{International Conference on Language Resources and
  Evaluation}, pages 3483--3487. European Language Resources Association.

\bibitem[{Gries(2013)}]{Gries2013}
St.~Th Gries. 2013.
\newblock \href {https://doi.org/10.1075/ijcl.18.1.09gri} {{50-something years
  of work on collocations: What is or should be next}}.
\newblock \emph{International Journal of Corpus Linguistics}, 18(1):137--165.

\bibitem[{Gr{\"{u}}nwald and Rissanen(2007)}]{GrunwaldPandRissanen2007}
P.~Gr{\"{u}}nwald and J.~Rissanen. 2007.
\newblock \emph{{The Minimum Description Length Principle}}.
\newblock MIT Press.

\bibitem[{Langacker(2008)}]{l08}
R.~Langacker. 2008.
\newblock \emph{{Cognitive Grammar A basic introduction}}.
\newblock Oxford University Press, Oxford.

\bibitem[{Matusevych et~al.(2013)Matusevych, Alishahi, and
  Backus}]{Matusevych2013}
Y.~Matusevych, A.~Alishahi, and A.~Backus. 2013.
\newblock \href {http://www.aclweb.org/anthology/W13-2606} {{Computational
  Simulations of Second Language Construction Learning}}.
\newblock In \emph{Proceedings of the Workshop on Cognitive Modeling and
  Computational Linguistics}, pages 47--56. Association for Computational
  Linguistics.

\bibitem[{Nguyen et~al.(2016)Nguyen, Nguyen, Pham, and Pham}]{nn}
Dat Quoca Dai~Quocb Nguyen, Dat Quoca Dai~Quocb Nguyen, Dang~Ducc Pham, and
  Son~Baod Pham. 2016.
\newblock \href {https://arxiv.org/pdf/1412.4021.pdf} {{A robust
  transformation-based learning approach using ripple down rules for
  part-of-speech tagging}}.
\newblock \emph{AI Communications}, 29(3):409--422.

\bibitem[{Petrov et~al.(2012)Petrov, Das, and McDonald}]{pdm12}
S.~Petrov, D.~Das, and R.~McDonald. 2012.
\newblock \href {https://www.aclweb.org/anthology/L12-1115/} {{A universal
  part-of-speech tagset}}.
\newblock In \emph{Proceedings of the Eighth Conference on Language Resources
  and Evaluation}, pages 2089--2096. European Language Resources Association.

\bibitem[{Sampson(2002)}]{Sampson2002}
G.~Sampson. 2002.
\newblock \href {https://doi.org/10.1515/tlir.19.1-2.73} {{Exploring the
  richness of the stimulus}}.
\newblock \emph{The Linguistic Review}, 19(1-2):73--104.

\bibitem[{Sardinha(2018)}]{Sardinha2018}
T.~Sardinha. 2018.
\newblock \href {https://doi.org/10.1075/ijcl.15026.ber} {{Dimensions of
  variation across Internet registers}}.
\newblock \emph{International Journal of Corpus Linguistics}, 23(2):125--157.

\bibitem[{Tayyar~Madabushi et~al.(2020)Tayyar~Madabushi, Romain, Divjak, and
  Milin}]{tayyar-madabushi-etal-2020-cxgbert}
H.~Tayyar~Madabushi, L.~Romain, D.~Divjak, and P.~Milin. 2020.
\newblock \href {https://doi.org/10.18653/v1/2020.coling-main.355}
  {{C}x{GBERT}: {BERT} meets construction grammar}.
\newblock In \emph{Proceedings of the 28th International Conference on
  Computational Linguistics}, pages 4020--4032. International Committee on
  Computational Linguistics.

\bibitem[{Tiedemann(2012)}]{Tiedemann2012}
J.~Tiedemann. 2012.
\newblock \href {https://www.aclweb.org/anthology/L12-1246/} {{Parallel Data,
  Tools and Interfaces in OPUS}}.
\newblock In \emph{Proceedings of the International Conference on Language
  Resources and Evaluation}. European Language Resources Association.

\bibitem[{Wible and Tsao(2010)}]{Wible2010}
D.~Wible and N.~Tsao. 2010.
\newblock \href {https://www.aclweb.org/anthology/W10-0804} {{StringNet as a
  Computational Resource for Discovering and Investigating Linguistic
  Constructions}}.
\newblock In \emph{Proceedings of the Workshop on Extracting and Using
  Constructions in Computational Linguistics}, pages 25--31. Association for
  Computational Linguistics.

\end{thebibliography}
\bibliographystyle{acl_natbib}

\end{document}